\documentclass[11pt,letterpaper,logo,twocolumn]{style}

\usepackage[numbers]{natbib}
\usepackage{graphicx}
\usepackage{booktabs}
\usepackage{amsmath,amsfonts,amssymb}
\usepackage{subcaption}
\usepackage{multirow}
\usepackage{colortbl}
\usepackage{listings}
\usepackage{xparse}
\usepackage{fontawesome5}
\usepackage{threeparttable}

\graphicspath{{./}{fig/}{figures/}{plot/}{pdf/}{table/}}
\usepackage{amsthm}
\tcbuselibrary{skins,breakable}
\tcbuselibrary{listingsutf8}
\usepackage{titletoc}
\usepackage{pifont}
\usepackage{mathtools}
\usepackage{bbm}
\usepackage{makecell}
\usepackage{adjustbox}

\usepackage{algorithm}
\usepackage{algpseudocode}
\usepackage{caption}

\usepackage{multirow}
\usepackage{makecell}
\usepackage{multicol}
\usepackage{multirow}
\usepackage{url}
\usepackage{amsmath,amsfonts,amsthm} 

\usepackage{balance}
\usepackage{color}
\usepackage{tabularx}
\usepackage{array}
\usepackage{booktabs}
\usepackage{amssymb}
\usepackage{bbding}
\usepackage{hyperref}
\usepackage{xcolor} 
\definecolor{tsneCaptionRed}{HTML}{D93D3A}
\definecolor{tsneCaptionBlue}{HTML}{317FBC}
\definecolor{tsneCaptionGreen}{HTML}{9BB49D}
\usepackage{booktabs} 
\usepackage{pifont}
\usepackage{colortbl}
\usepackage{enumitem}
\usepackage{bm}
\usepackage{subcaption}
\usepackage{tikz}
\usetikzlibrary{shapes}
\newcommand{\revised}[1]{#1}
\newenvironment{paragraphrevised}{}{}

\newcommand\blfootnote[1]{%
\begingroup
\renewcommand\thefootnote{}\footnote{#1}%
\addtocounter{footnote}{-1}%
\endgroup
}

\title{Parallel Test-Time Scaling for Latent Reasoning Models}
\runningtitle{Parallel Test-Time Scaling for Latent Reasoning Models}

\PublicDate{Accepted at ACL 2026 Main Conference}

\author{%
  {\Authfont
    \textbf{Runyang You}\textsuperscript{1} \quad
    \textbf{Yongqi Li}\textsuperscript{1}\advisor \quad
    \textbf{Meng Liu}\textsuperscript{2} \quad
    \textbf{Wenjie Wang}\textsuperscript{3} \quad
    \textbf{Liqiang Nie} \textsuperscript{4} \quad
    \textbf{Wenjie Li}\textsuperscript{1} \quad
  }\\
  {\Affilfont
    \textsuperscript{1} Hong Kong Polytechnic University, \quad
    \textsuperscript{2} Shandong Jianzhu University \\
    \textsuperscript{3} University of Science and Technology of China
    \quad
    \textsuperscript{4} Harbin Institute of Technology (Shenzhen) \\
    \texttt{runyang.y@outlook.com, liyongqi0@gmail.com}
  }
}

\begin{document}
\blfootnote{$^\dagger$Corresponding author.}

\begin{abstract}

Parallel test-time scaling (TTS) is a pivotal approach for enhancing large language models (LLMs), typically by sampling multiple token-based chains-of-thought in parallel and aggregating outcomes through voting or search.
Recent advances in latent reasoning, where intermediate reasoning unfolds in continuous vector spaces, offer a more efficient alternative to explicit Chain-of-Thought, yet whether such latent models can similarly benefit from parallel TTS remains open, mainly due to the absence of sampling mechanisms in continuous space, and the lack of probabilistic signals for advanced trajectory aggregation.
\
This work enables parallel TTS for latent reasoning models by addressing the above issues. For sampling, we introduce two uncertainty-inspired stochastic strategies: Monte Carlo Dropout and Additive Gaussian Noise. For aggregation, we design a Latent Reward Model (LatentRM) trained with step-wise contrastive objective to score and guide latent reasoning. Extensive experiments and visualization analyses show that both sampling strategies scale effectively with compute and exhibit distinct exploration dynamics, while LatentRM enables effective trajectory selection. Together, our explorations open a new direction for scalable inference in continuous spaces.


\end{abstract}

\newcommand{\TitleLinks}{%
\centering
    \vspace{6pt}
    {\noindent\absfont\fontsize{11}{13}\selectfont
    \faGithub\ Project Page: \url{https://github.com/ModalityDance/LatentTTS}\par}%
}

\begin{teaserfigure}
    \centering
        \begin{subfigure}[t]{0.31\textwidth}
        \includegraphics[width=\textwidth]{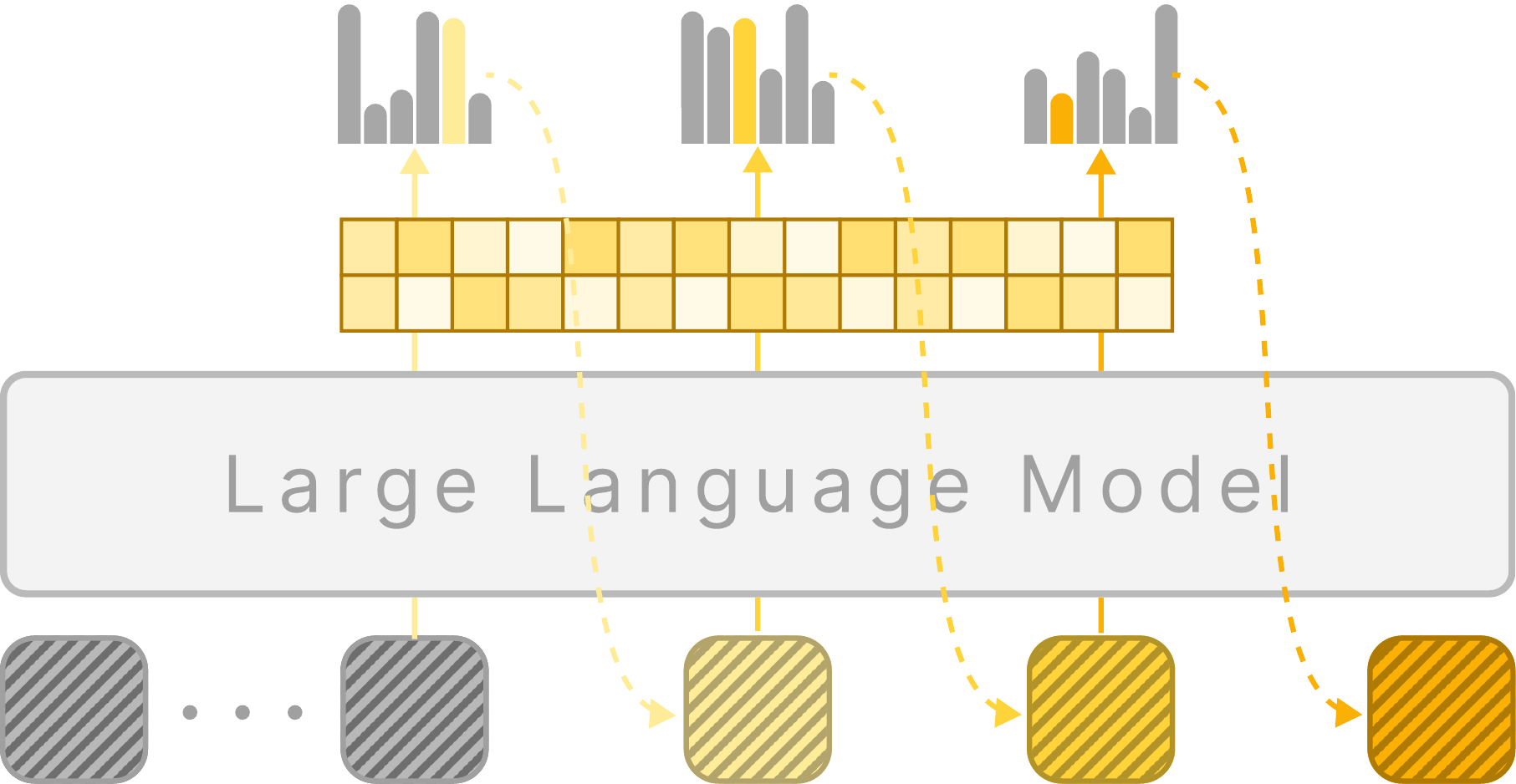}
        \caption{Token-based Sampling}
        \label{fig:main-token}
    \end{subfigure}
    \hfill
    \begin{subfigure}[t]{0.31\textwidth}
        \includegraphics[width=\textwidth]{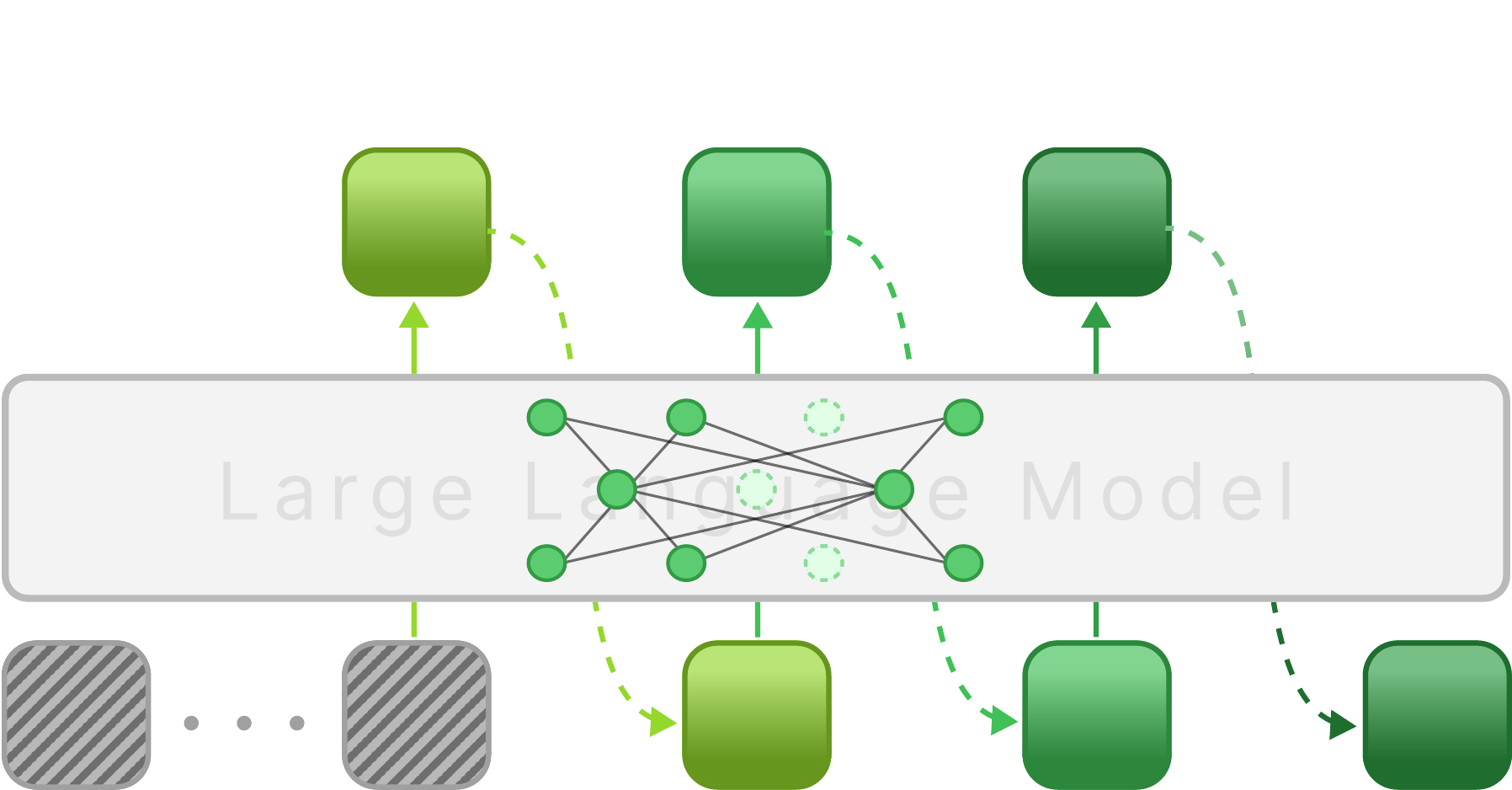}
        \caption{Monte Carlo Dropout}
        \label{fig:main-latent+dropout}
    \end{subfigure}
    \hfill
    \begin{subfigure}[t]{0.31\textwidth}
        \includegraphics[width=\textwidth]{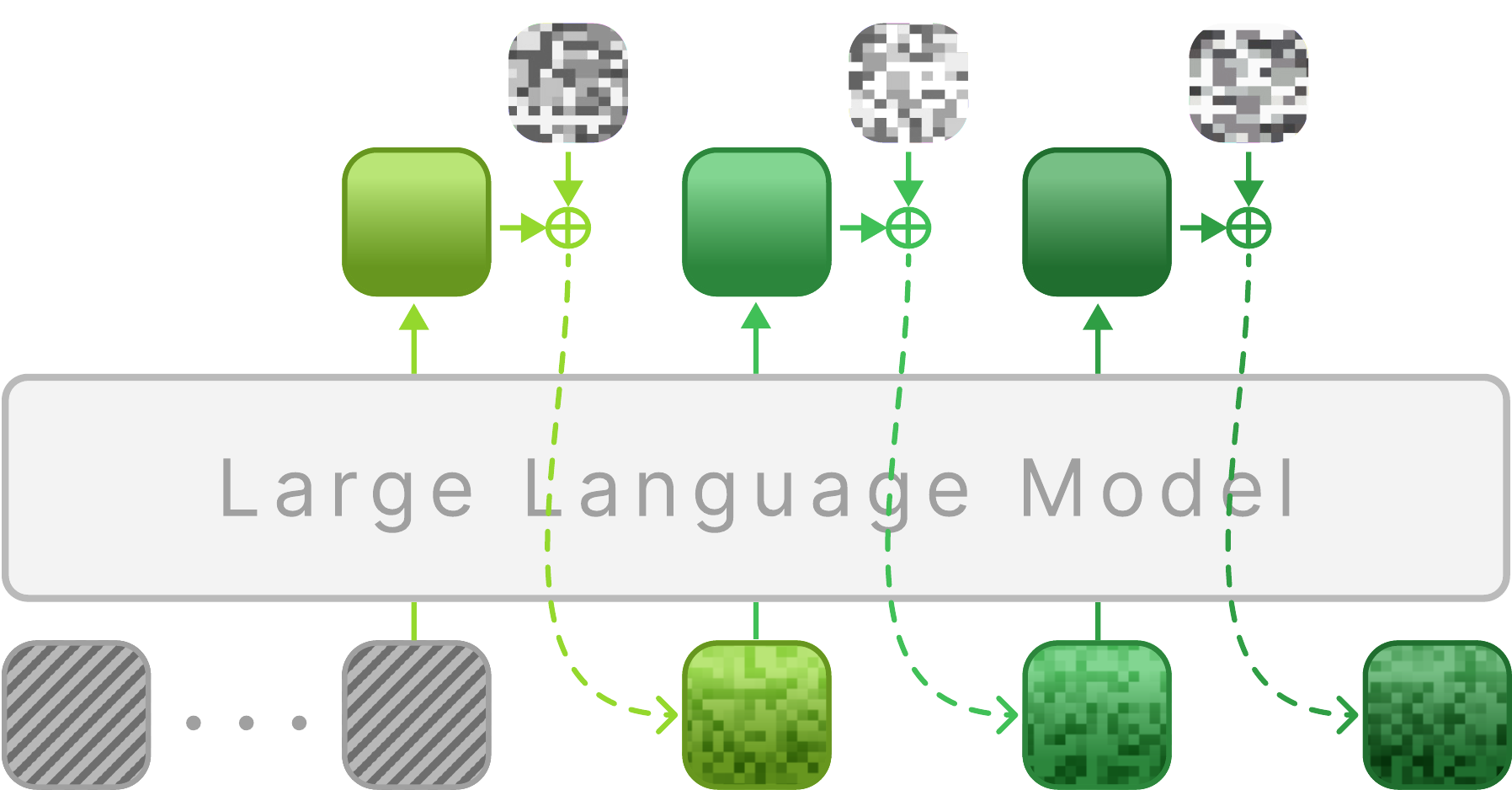}
        \caption{Additive Gaussian Noise}
        \label{fig:main-latent+noise}
    \end{subfigure}
    \caption{Sampling mechanisms for token-based generation (\ref{fig:main-token}) and our proposed approaches for the latent setting (\ref{fig:main-latent+dropout} \& \ref{fig:main-latent+noise}).
(\ref{fig:main-token}): Multinomial sampling over token probabilities.
(\ref{fig:main-latent+dropout}): Monte Carlo Dropout (MC-dropout): Inference via randomly sampled dropout masks.
(\ref{fig:main-latent+noise}): Additive Gaussian Noise (AGN): independent Gaussian perturbations.
}
    \label{fig:main}
\end{teaserfigure}

\maketitle

\section{Introduction} \label{sec:intro}

Large language models (LLMs) have achieved remarkable performance on challenging tasks through test-time scaling (TTS), where more inference compute leads to better results.
Such scaling is often realized through explicit Chain-of-Thought (CoT) reasoning~\cite{wei_chain--thought_2022}, where models verbalize intermediate solving steps in natural language, generate long token sequences over which compute can be scaled.
\
One promising direction is to scale in parallel~\cite{fu_deep_2025, qu_optimizing_2025,snell_scaling_2024}, which samples multiple reasoning paths and aggregates outcomes via methods such as majority voting~\cite{wang_self-consistency_2022}, best-of-$N$ ~\cite{zhou_least--most_2022}, or guided search~\cite{uesato_solving_2022}. 
Through parallel scaling, models transform extra inference compute directly into stronger capabilities, without any need for parameter updates or retraining.

Moving beyond explicit token-by-token reasoning, recent works show that reasoning can instead unfold in latent space, with vector representations replacing tokens as the intermediate steps in autoregressive generation~\cite{li_implicit_2025, chen_reasoning_2025}. This paradigm, also known as continuous CoT (CCOT), has the potential to match or even surpass explicit CoT while being more compact and computationally efficient, resembling intuitive human reasoning~\cite{COCONUT, shen_codi_2025, COLAR, wu_parallel_2025}. 
By effectively compressing or distilling explicit reasoning into continuous representations, these methods not only speed up inference but also capture abstract patterns difficult to express in natural language, positioning latent reasoning as a promising direction for advancing LLM capabilities.

Given the promising potential of latent reasoning, and that parallel TTS can effectively utilize additional generated tokens to deliver superior  performance~\cite{wang_self-consistency_2022, muennighoffS1SimpleTesttime2025, snell_scaling_2024}, a natural question arises: \textbf{can latent reasoning models also benefit from parallel TTS?}
Extending parallel TTS into latent reasoning models is appealing but non-trivial.
First, latent reasoning models lack the fundamental sampling capability:
while token-based models generate logits that enable commonly-used sampling strategies like top-k or nucleus sampling--as illustrated in Figure~\ref{fig:main-token}, latent models operate on continuous vectors without an explicit probability distribution and therefore \revised{lack an inherent mechanism to stochastically generate multiple reasoning paths}.

Second, the aggregation mechanism used in token-based models does not directly apply to latent reasoning models: while token-based methods typically utilize token-level probabilities to rank reasoning trajectories, latent reasoning models provide no inherent likelihoods or stepwise scores, making it difficult to evaluate or aggregate the sampled latent reasoning paths.

To address these challenges, we re-think both sampling and aggregation for latent reasoning, proposing effective solutions tailored to the continuous setting.
For sampling, to ensure informativeness and controllability of the sampling space, we draw on uncertainty estimation theory~\cite{gawlikowski_survey_2023, NEURIPS2019_1cc8a8ea} and propose two simple yet effective strategies to introduce stochasticity into latent reasoning: Monte Carlo dropout (MC-dropout) to capture epistemic uncertainty; and Additive Gaussian Noise (AGN) to simulate aleatoric uncertainty, as illustrated in Figure~\ref{fig:main}.
\ 
For aggregation, we propose the Latent Reward Model (LatentRM), a dedicated scorer that evaluates and guides the progression of latent reasoning at each intermediate step.
LatentRM is trained via a step-wise contrastive objective that discriminates among candidate thoughts at each reasoning step, enabling fine-grained, position-sensitive scoring.

Building on these designs, extensive experiments and visualization analyses reveal that both proposed sampling strategies not only scale effectively with increased compute but also exhibit distinct exploration dynamics in latent space: 
MC-Dropout promotes structured, directed expansion toward unconventional solutions, resulting in higher coverage, whereas Additive Gaussian Noise drives broad and isotropic exploration that enriches diversity around the deterministic center.
\ %
For aggregation, LatentRM enables consistent gains with best-of-$N$ and beam search across compute budgets. Ablation studies confirm that contrastive supervision and stochastic rollouts are both crucial.
Overall, 
our findings demonstrate that parallel TTS can transfer to latent reasoning models through redesigned sampling and aggregation, opening a new pathway for scalable inference in latent space.

Collectively, core contributions are as follows:
\begin{itemize}
    \item We introduce parallel test-time scaling into latent reasoning models, enabling a key scaling capability that was previously exclusive to token-based reasoning paradigms.
    \item We address the fundamental challenge of sampling in continuous latent space by proposing two complementary strategies—Monte Carlo Dropout and Additive Gaussian Noise to enable controlled and informative stochastic latent reasoning.
    \item We design the Latent Reward Model (LatentRM), a dedicated scorer under step-wise contrastive supervision to evaluate and guide latent reasoning, enabling effective aggregation in the latent setting.
\end{itemize}

\section{Related Work}
\subsection{Test-Time Scaling}
Large language models increasingly rely on test-time scaling (TTS)—allocating more computation at inference to improve reasoning quality. Two main axes have emerged. The sequential axis advances by generating longer reasoning traces~\cite{muennighoffS1SimpleTesttime2025,wang_scaling_2025}, a strategy employed in recent reasoning-oriented models~\cite{deepseek-ai_deepseek-v3_2024,yang_qwen3_2025,you_textr2textec_2025,fu_deep_2025,lee_evolving_2025}. The parallel axis generates multiple reasoning trajectories and aggregates them, as in Self-Consistency~\cite{wang_self-consistency_2022}, Best-of-N~\cite{zhou_least--most_2022}, and tree-structured exploration methods~\cite{uesato_solving_2022}. These two axes are often combined, enabling flexible accuracy–compute trade-offs under deployment constraints.
Recent works have sought to improve the efficiency of TTS primarily by refining aggregation strategies and decision rules, such as adaptive voting and confidence-based stopping criteria~\cite{snell_scaling_2024,brown_large_2024}. 
Yet all of these approaches ultimately rest on token-level sampling, decoding by drawing tokens from the predictive distribution. Such mechanism is intrinsic to token-based reasoning, but is fundamentally incompatible with latent reasoning, which operates over continuous representations.

\subsection{Latent Reasoning}

Chain-of-Thought (CoT) reasoning has proven highly effective, yet it remains constrained by its reliance on natural language~\cite{zhu_survey_2025}. This introduces inefficiency--most tokens add little value and fail to capture cognition like abstract insights or intuitive leaps, making step-by-step verbalization an unnatural limit. These limitations motivate the shift toward latent reasoning~\citep{chen_emergence_2025,chen_reasoning_2025}.
One line of work pursues architectural modifications, such as dynamically skipping or repeating transformer layers, or introducing auxiliary modules to adjust computational cost~\cite{chen_inner_2025,geiping_scaling_2025,mohtashami_cotformer_2024}.
More recently, research has shifted to latent autoregressive reasoning, where models build trajectories directly in hidden space.
COCONUT~\cite{COCONUT} pioneered latent generation using the last hidden state as the next thought, training via curriculum learning, while CODI~\cite{shen_codi_2025} introduced latent auto-regression through self-distillation. While CoLaR~\cite{COLAR} and SoftCoT++~\cite{Softcot++} made initial attempts to inject noise for stochastic latent reasoning, these efforts remain preliminary.


\section{Preliminaries}

We begin by formalizing latent reasoning and outlining the inference framework that underlies our test-time scaling approach.

\paragraph{Latent Reasoning.}
A latent reasoning model performs autoregressive generation in continuous hidden space.
At each step, it produces a latent vector—the last hidden state from the transformer backbone—that represents an intermediate reasoning step.
This formulation bypasses token-level verbalization, providing a more compact and efficient alternative to standard explicit CoT reasoning.

\paragraph{Inference Process.}
Let the input sequence be $\bm{x} = [x_1, x_2, \ldots]$, the latent reasoning trajectory as
$\bm{h}_{1:T} = [\bm{h}_1, \bm{h}_2, \ldots, \bm{h}_T] \in \mathbb{R}^{T \times d}$
, where each $\bm{h}_t \in \mathbb{R}^d$ represents the hidden state (thought) at step $t$.
At inference time, the model generates the next latent thought autoregressively:
\[
{\bm{h}}_{t+1} = f_{\bm{\theta}}({\bm{h}}_{1:t}, \bm{x}),
\]
where \(f_{\bm{\theta}}\) is the LLM transformer blocks parameterized by \(\bm{\theta}\).
This process ends with the end-of-thinking token (\texttt{<|EoT|>}), which triggers the transition to explicit token generation for the final answer.
\paragraph{Reasoning Length Control.}
Different latent reasoning models control the reasoning length in distinct ways.
COCONUT~\cite{COCONUT} and CODI~\cite{shen_codi_2025} use predetermined sequence lengths, where a fixed 
$T$ is set and the \texttt{<|EoT|>} is manually inserted.
CoLaR~\cite{COLAR} introduces dynamic latent compression: the reasoning length (or “thinking speed”) is adjusted by a compression factor specified in the prompt (e.g., “thinking speed = 5×”), allowing the model to adaptively decide when to emit the EoT token and terminate reasoning.

\section{Method}

Building on the latent reasoning formulation above, we now introduce our test-time scaling framework, which consists of two key components: stochastic sampling and latent trajectory aggregation.

\subsection{Stochastic Sampling in Latent Space}
\label{sec:sampling}
To scale latent reasoning at test time, stochasticity must be introduced into the latent generation process, enabling the model to produce a diverse set of trajectories $\{\bm{h}^{(n)}\}_{n=1}^N$, where each 
    $\bm{h}^{(n)} = [\bm{h}^{(n)}_1,\ldots,\bm{h}^{(n)}_T]$ represents a sampled sequence of latent thoughts.
However, arbitrary noise can easily distort reasoning process.
To ensure that sampling is both controllable and meaningful,
we draw on uncertainty estimation theory,
which delineates two sources of uncertainty that provide principled probabilistic spaces for sampling: (1) \emph{epistemic uncertainty}, reflecting variability due to the model’s limited knowledge, and (2) \emph{aleatoric uncertainty}, arising from noise or ambiguity inherent in the inputs. 
Accordingly,
we propose two complementary strategies: \textbf{Monte Carlo dropout} (MC-dropout) to capture epistemic uncertainty, and \textbf{Additive Gaussian Noise} (AGN) to simulate aleatoric uncertainty. An overview of the two mechanisms is shown in Figure~\ref{fig:main-latent+dropout} and Figure~\ref{fig:main-latent+noise}.

We now detail the sampling procedures. Formal derivations and algorithmic details are provided in Appendix~\ref{sec:appendix-sampling} and Appendix~\ref{sec:appendix-sampling-algo}, respectively.

\paragraph{MC-dropout.}
We keep dropout active at inference with rate $p$, sampling binary masks $m^{(n)} \sim \text{Bernoulli}(p)$ applied to model weights $\bm{\theta}^{(n)}$:
\[
\bm{h}_{t+1}^{(n)} = f_{\bm{\theta}^{(n)}}({\bm{h}}_{1: t}^{(n)}, \bm{x}).
\]
Dropout is applied after the feed-forward layer in each Transformer block~\citep{gao_simcse_2021}, capturing \emph{epistemic uncertainty} as each pass samples a different weight configuration.

\paragraph{AGN.}
As a complementary sampling strategy, we add isotropic Gaussian noise directly to the latent thoughts to induce controlled stochasticity, as illustrated in Figure~\ref{fig:main-latent+noise}.
Specifically, a random perturbation is drawn and applied at each reasoning step as
\[{\bm{\epsilon}}_t^{(n)} \sim \mathcal{N}(0,\sigma^2 \mathbf{I}), \quad
{\bm{h}}_{t}^{(n)*} = {\bm{h}}_{t}^{(n)} + {\bm{\epsilon}}_t^{(n)},
\]
where $\mathbf{I}$ is the identity matrix and $\boldsymbol{\epsilon}_t^{(n)}$ denotes zero-mean Gaussian noise with standard deviation~$\sigma$.
The model then continues autoregressive inference based on the perturbed trajectory:
\[
{\bm{h}}_{t+1}^{(n)} = f_{\bm{\theta}} \left({\bm{h}}_{1:t}^{(n)*}, \bm{x}\right).
\]
This procedure models \emph{aleatoric uncertainty}, as the variance is controlled solely by the noise scale~$\sigma$, independent of the model parameters.

\subsection{Latent Trajectories Aggregation}
\label{subsec:methods-aggregation}
A key challenge in extending parallel TTS to latent reasoning lies in aggregation.
Unlike token-based TTS, where log-likelihoods provide a natural scoring mechanism, latent trajectories are continuous vectors without explicit scores. This prevents direct application of best-of-$N$ or beam search.
While process reward models (PRMs) \cite{zhang_openprm_2024,math-shepherd} could be a natural choice for evaluating intermediate reasoning steps, latent thoughts are abstract vectors without linguistic form and cannot be interpreted or scored by existing PRMs.
To this end, we propose the \textbf{Latent Reward Model (LatentRM)}, a dedicated reward model for latent thoughts that enables effective aggregation strategies.

\paragraph{Architecture.}
We extend the latent reasoning backbone with an additional scoring head that maps hidden states to a scalar. LatentRM $g_{\bm{\phi}}$ takes in the input prompt $x$ and the generated latent trajectory up to step $t$ and outputs a score:
\[
r_t = g_{\bm{\phi}}(\bm{x}, \bm{h}_{1: t}) \in \mathbb{R},
\]
which estimates the promise of continuing from the current thought $\bm{h}_t$.

\paragraph{Inference.}
During inference, LatentRM evaluates each candidate trajectory by computing the sum of logits $\sum_{t} r_{t}$ over the generated sequence, serving as a proxy for the relative quality of thought $\bm{h}_{1:t}$. This logit-summing strategy is justified in Appendix~\ref{sec:appendix-scoring}, which shows that cumulative logits can solely determine trajectory ranking.

\paragraph{Data.}
To supervise the learning of LatentRM, we construct thought-level quality labels by estimating the empirical correctness of each intermediate thought.
Specifically, for each input $x$,
we sample $N$ trajectories $\bm{H}=\{\bm{h}^{(n)}\}_{n=1}^N$, and for every thought within every trajectory, we rollout $M$ stochastic completions, obtaining a set of final answers$\{a_{m}\}_{m=1}^M$, and compute
\[
\tilde{y} = \frac{1}{M} \sum_{m=1}^M \mathbb{I}\left\{a_{m} = a^* \right\},
\]
as a proxy for the quality of thought, where $a^*$ denotes the ground-truth answer and $\mathbb{I}\{\cdot\}$ as the indicator function.

\paragraph{Objective.}
A straightforward approach is to treat each thought independently and optimize LatentRM with binary cross-entropy (BCE) loss, which is commonly used for training PRMs.
However, this approach performs poorly in practice, as it provides only isolated supervision per candidate and lacks relative comparison across thoughts at the same step.
To address this, we adopt a step-wise contrastive formulation. At each step $t$, the scores of all $N$ candidates are compared via a softmax,
\[
p_t^{(n)} = \frac{\exp(r_t^{(n)})}{\sum_{n'=1}^N \exp(r_t^{(n')})}.
\]
LatentRM is then trained with the negative log-likelihood loss
\[
\mathcal{L} = -\sum_t \sum_{n=1}^N y_t^{(n)} \log p_t^{(n)}.
\]

\section{Experiment Setup}
\subsection{Benchmarks and Models.}
\label{subsec:exp-benchmarks-models}

\paragraph{Benchmarks.} 
Following previous works~\cite{shen_codi_2025, COLAR, COCONUT},
evaluation is conducted on three benchmarks:
(1) {GSM8K-Test}, the official test split of GSM8K (1,319 test samples)\footnote{\url{https://huggingface.co/datasets/gsm8k}};
(2) {GSM8K-Hard}~\cite{GSM-HARD}, a more challenging version of GSM8K-Test where numbers are scaled to larger magnitudes to increase problem difficulty with (1,319 test samples)\footnote{\url{https://huggingface.co/datasets/reasoning-machines/gsm-hard}},
and (3) {MultiArith}~\cite{uesato_solving_2022}, a dataset focusing on multi-step arithmetic reasoning (600 test samples)\footnote{\url{https://huggingface.co/datasets/lighteval/multi_arith}}.
\paragraph{Latent reasoning models/backbones.}
We evaluate five representative latent reasoning models: (1) {COCONUT} \citep{COCONUT}, which progressively replaces CoT steps with latent thoughts; (2) {CODI} \citep{shen_codi_2025}, which self-distills CoT into latent space; and (3) {CoLaR} \citep{COLAR}, which performs dynamic latent compression with reinforcement learning. 
All experiments leverage officially released checkpoints, where COCONUT and CODI are backboned on GPT-2 \revised{(124M)}, and CoLaR on Llama-3.2-1B. (4) {Latent-SFT} (Llama-3.2-1B-Instruct)~\cite{latent-sft_2025} and (5) {Render-of-Thought (RoT)} with Qwen3-VL at 2B and 4B parameter scales~\cite{renderofthought_2026}.
Following the default configurations in the original papers, for COCONUT and CODI, we fix the number of latent thoughts to $T=6$; for CoLaR, thinking speed set to $2$. For ColaR, Latent-SFT and RoT, we set the maximum number of latent thoughts to 64.

\subsection{Sampling Evaluation}
To evaluate the effectiveness of different stochastic sampling methods, we measure how well each method scales with the number of sampled trajectories. Specifically, \emph{coverage} quantifies the fraction of problems for which at least one of the $N$ sampled trajectories yields the correct answer:
\[
\text{coverage} = \frac{1}{|D|}\sum_{x\in D}\mathbb{I}\{\exists n\le N:{a}^{(n)}=a^*\},
\]
This metric is equivalent to \emph{pass@k} when sampling size $N$ and cutoff $k$ are identical~\cite{zhang_openprm_2024,brown_large_2024}. 
Higher coverage indicates stronger sampling effectiveness—i.e., a greater ability to uncover correct reasoning paths as the inference budget increases.
All sampling schemes are evaluated under equal $N$ for fair comparison.

\subsection{Aggregation Evaluation}
\label{subsec:exp-aggregation}
We evaluate the proposed LatentRM through two of its supported aggregation strategies:
(1) \emph{best-of-$N$} selection scored by LatentRM,
(2) \emph{beam search} guided by LatentRM,
compared against \emph{majority voting} as a non-parametric baseline. \revised{We adopt MC-dropout with $p=0.2$ as the default sampling setting.}
To ensure fairness, we match all methods under the same compute budget: best-of-$N$ and majority voting use $N$ independent samples, while beam search adopts a beam size of
$\sqrt{N}$ for comparable decoding cost~\cite{zhang_openprm_2024, snell_scaling_2024}. 
Training configurations and optimization details for LatentRM are provided in Appendix~\ref{sec:appendix-training-latentRM}, and full inference procedures for aggregation strategies are described in Appendix~\ref{sec:appendix-aggregation-algo}.

\begin{figure}[t]
    \centering
    \begin{subfigure}[b]{\linewidth}
        \includegraphics[width=\linewidth]{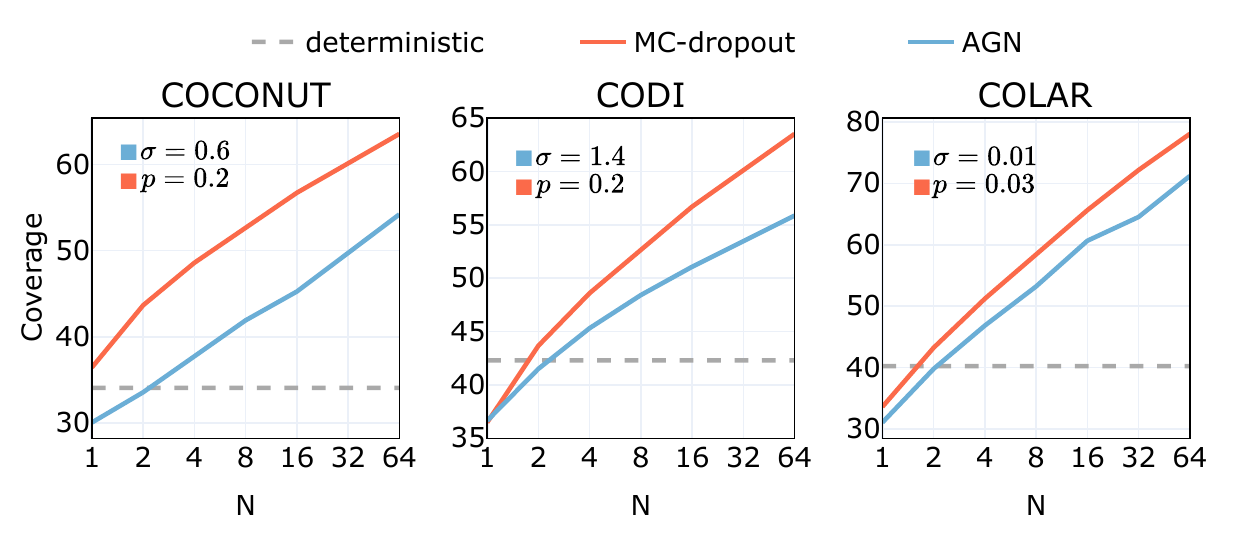}
        \vspace{-2em}
        \caption{GSM-Test}
        \label{fig:sampling:gsm-test}
    \end{subfigure}
    \begin{subfigure}[b]{\linewidth}
        \includegraphics[width=\linewidth]{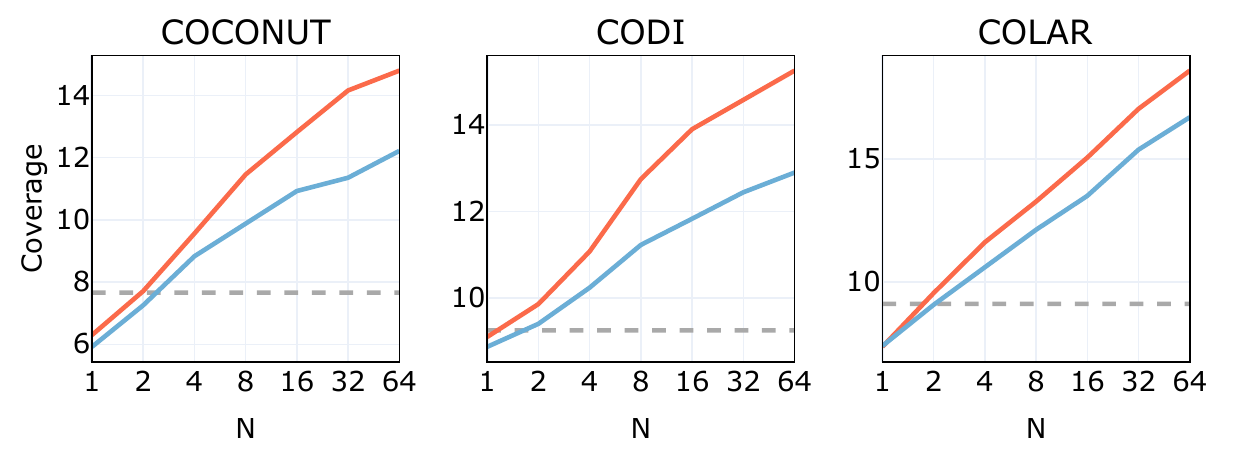}
        \vspace{-2em}
        \caption{GSM-Hard}
        \label{fig:sampling:gsm-hard}
    \end{subfigure}
    \begin{subfigure}[b]{\linewidth}
        \includegraphics[width=\linewidth]{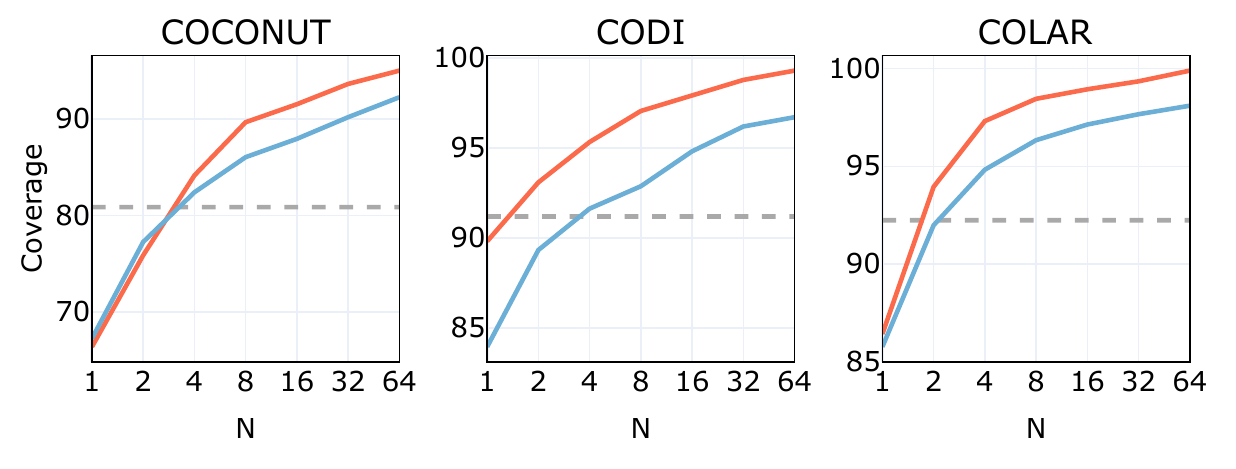}
        \vspace{-2em}
        \caption{MultiArith}
        \label{fig:sampling:gsm-multiarith}
    \end{subfigure}
    \caption{\revised{Coverage (\%) versus N plot for COCONUT, CODI, and CoLaR on GSM-Test (\ref{fig:sampling:gsm-test}), GSM-Hard (\ref{fig:sampling:gsm-hard}) and MultiArith (\ref{fig:sampling:gsm-multiarith}). Each subplot compares MC-dropout and AGN using the optimal hyperparameter. Higher coverage indicates a larger fraction of problems solved by $N$ attempts. Results are reported as the mean over three runs.}}
    \label{fig:sampling}
\end{figure}
\section{
Results of Sampling
}

\subsection{Main Results}
\label{subsec:sampling-main-results}

We systematically evaluate both stochastic sampling strategies by varying the sample count (N) and measuring solution coverage. We tune the AGN hyperparameter $\sigma$ over $[0.01, 1.5]$ and the MC-Dropout probability $p$ over $[0, 1]$, using binary search within each interval to maximize coverage@$64$. We then plot two curves using the optimal hyperparameters for each method. Additional results on harder benchmarks are provided in Appendix~\ref{sec:appendix-harder-benchmarks}.
For deployment without exhaustive sweeps, we use the following \textbf{heuristic ranges} informed by backbone architecture: (i) GPT-2--style models (e.g., COCONUT, CODI) work well with dropout rate $p \in [0.1, 0.3]$ and noise scale $\sigma \in [0.5, 0.7]$; (ii) Llama-3.2-1B--style models (e.g., CoLaR, Latent-SFT) are more stable with smaller perturbations, $p \in [0.01, 0.03]$ and $\sigma \in [0.01, 0.03]$.
In practice, fixing $(p,\sigma)$ to a single pair within these bands and using it across benchmarks yields coverage within a negligible gap of values obtained by per-dataset tuning, so practitioners need not re-tune for each split.
Figure~\ref{fig:sampling} presents four key findings: 
(1) both MC-dropout and AGN can effectively scale, as coverage increases monotonically with larger sample sizes; 
(2) the marginal improvement diminishes as $N$ grows, suggesting a saturation effect where additional samples contribute less to coverage gains;
(3) increased sampling narrows inter-model performance gaps. Notably, at N=64, COCONUT and CODI achieve nearly equivalent coverage, despite CODI's clear superiority at N=1.
(4) MC-dropout consistently achieves higher coverage across nearly all $N$, highlighting its advantage as a more reliable stochastic sampling approach.

\begin{figure}[t]
  \centering
  \includegraphics[width=1.\linewidth]{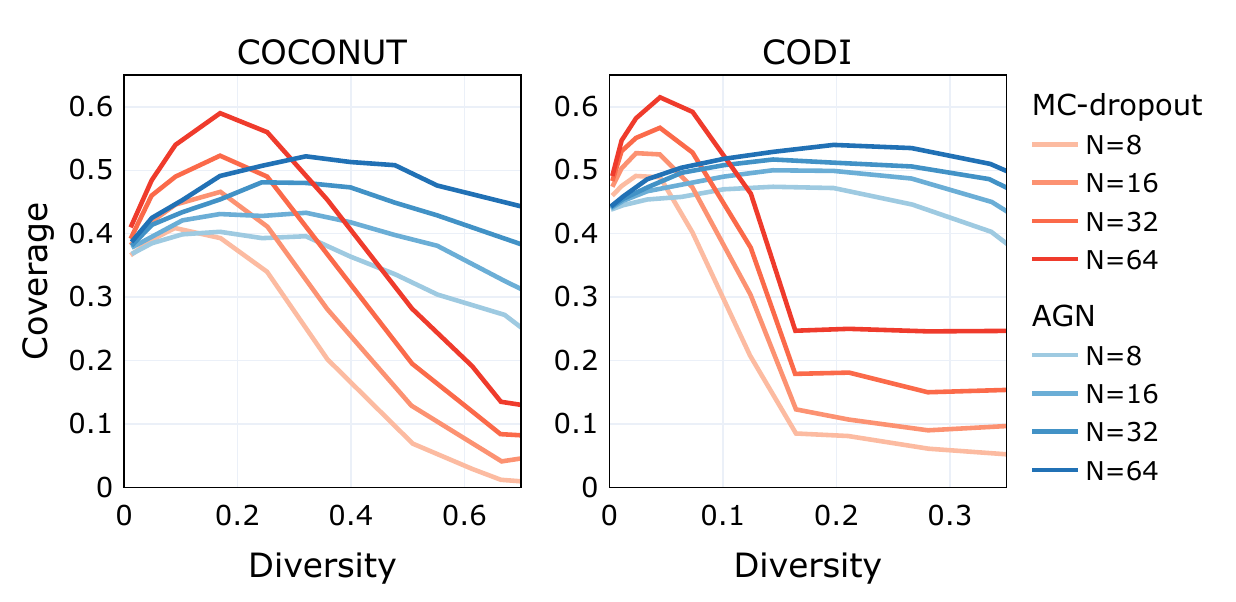}
  \caption {
Coverage versus diversity for MC-dropout (red) and AGN (blue) with $N \in \{4, 8, 16, 32\}$ by sweeping $p$ and $\sigma$ to span a range of diversity values. Darker shades indicate larger $N$. Results are shown for COCONUT (left) and CODI (right) on GSM-Test.
  }
  \label{fig:div_cov}
\end{figure}
\begin{figure}[t]
  \centering
  \includegraphics[width=1.\linewidth]{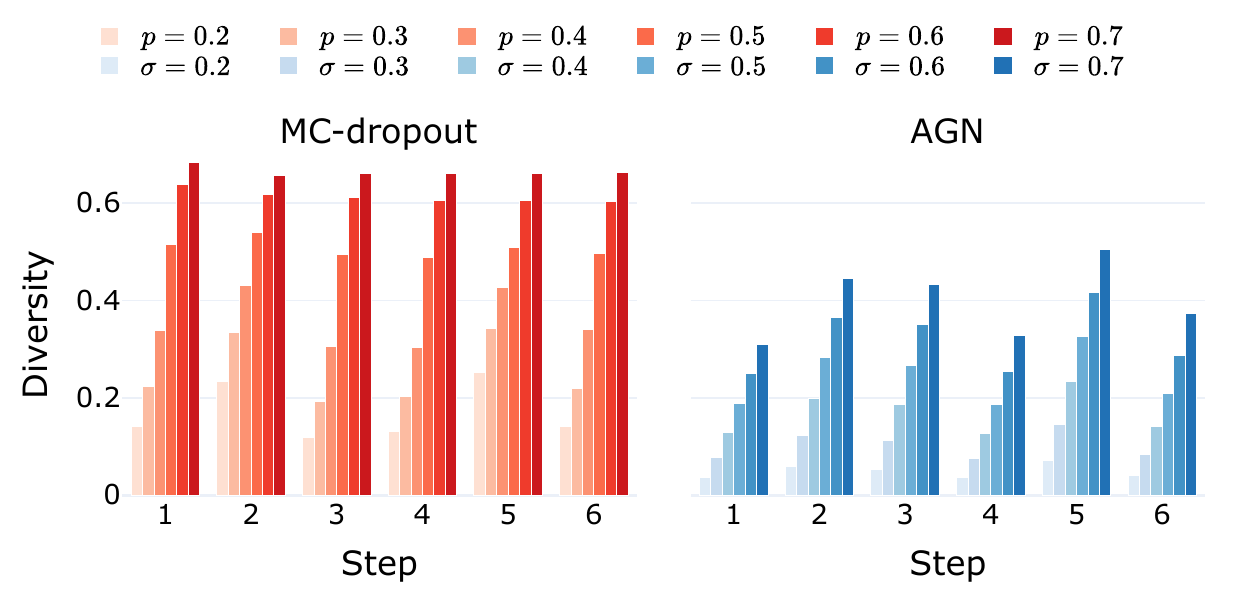}
  \caption {
Diversity of latent trajectories across reasoning steps on GSM-Test with COCONUT. Left: MC-dropout ($p = 0.1$ - $0.5$). Right: AGN ($\sigma = 0.1$ - $0.5$).
}
\label{fig:diversity}
\end{figure}

\subsection{Generalization to Larger Backbones}
\begin{table}[t]
\centering
\small
\vspace{0.5em}
\resizebox{\linewidth}{!}{
\begin{tabular}{ll ccc}
\toprule
\textbf{Backbone} & \textbf{Benchmark} & \textbf{Det.} & \textbf{Cov@8} & \textbf{Cov@16} \\
\midrule
\multirow{2}{*}{\shortstack[l]{\textbf{Latent-SFT} \\ \scriptsize{(Llama-3.2-1B)}}} 
    & GSM8K & 0.445 & 0.585 & 0.649 \\
    & MultiArith & 0.934 & 0.962 & 0.967 \\ \midrule
\multirow{3}{*}{\shortstack[l]{\textbf{RoT-4B} \\ \scriptsize{(Qwen3-VL)}}} 
    & GSM8K & 0.375 & 0.394 & 0.397 \\
    & GSM8K-Hard & 0.139 & 0.143 & 0.143 \\
    & {MATH500} & 0.203 & 0.218 & 0.220 \\ \midrule
\multirow{3}{*}{\shortstack[l]{\textbf{RoT-2B} \\ \scriptsize{(Qwen3-VL)}}} 
    & GSM8K & 0.233 & 0.257 & 0.257 \\
    & GSM8K-Hard & 0.086 & 0.091 & 0.092 \\
    & {MATH500} & 0.115 & 0.128 & 0.130 \\
\bottomrule
\end{tabular}
}
\caption{Generalization across larger backbones and scales. $N$ denotes the number of sampled latent trajectories. MC-dropout is adopted as the default setting.}
\label{tab:generalization}
\end{table}
    
To evaluate the architectural robustness and scalability of our framework, we extend our experiments to more capable backbones, including {Latent-SFT}~\cite{latent-sft_2025} and {Render-of-Thought (RoT)}~\cite{renderofthought_2026}, with parameter counts up to 4B. As shown in Table~\ref{tab:generalization}, our parallel TTS framework yields consistent performance improvements across various architectures and scales. In particular, on the challenging {MATH500} benchmark, RoT-4B improves from 20.3\% to 22.0\% with $N=16$, highlighting the framework's ability to generalize to larger models and complex mathematical reasoning tasks.

\subsection{Analysis on Diversity}
The key question to ask is: \emph{what makes a good sampling strategy for latent reasoning?} 
Beyond simply scaling the sampling budget $N$ to boost problem-solving success, a good sampling strategy should encourage diverse reasoning trajectories rather than expending computational effort on redundant or overly similar paths. Importantly, this pursuit of diversity must not come at the cost of overall coverage.

\paragraph{Definition.}
To formalize this intuition, we introduce the notion of sampling  \emph{diversity} as the average pairwise cosine dissimilarity among latent thoughts across reasoning steps:
$$
\text{diversity} = \frac{1}{|D|T}\sum_{x\in D}\sum_{t=1}^{T} d_t(x),
$$
\[\centering
\resizebox{0.99\linewidth}{!}{
where
$
d_t(x) = \frac{2}{N(N-1)}
\sum_{i<j}
(1 - \cos({\bm{h}}^{(i)}_{t}, {\bm{h}}^{(j)}_{t})), 
$}
\]
measures the dissimilarity at step $t$.
A higher diversity score reflects greater variability in the reasoning paths taken, suggesting a richer exploration of the problem space.

\paragraph{Diversity vs. coverage.}
To understand the interplay between diversity and performance, we plot coverage against diversity across different sampling method, while sweeping $p$ and $\sigma$ under varying $N$. The resulting trade-off curves are visualized in Figure~\ref{fig:div_cov}.
Key insights are as follow.
(1) Across models and methods, a clear “sweet spot” emerges, with coverage peaking at moderate diversity. This suggests that while some level of stochasticity is beneficial, excessive or insufficient diversity can hinder performance; and
(2) at larger diversity levels, AGN tends to maintain or even improve coverage, whereas MC-dropout shows a sharp decline. This highlights AGN's superior ability to preserve solution quality even when the injected randomness is high, making it a more robust choice for high-diversity exploration.
Overall, if high-diversity exploration is desired, AGN serves as a more reliable choice.

\paragraph{Step-wise dynamics.}
To examine how sampling affects reasoning over time, we analyze diversity at each step. As shown in Figure~\ref{fig:diversity}, MC-dropout maintains similar diversity across steps for a given $p$, reflecting its adaptivity: stochasticity follows the model’s own uncertainty, enabling consistent exploration. In contrast, AGN shows pronounced fluctuations because a fixed $\sigma$ perturbs latent vectors of varying scales unevenly, yielding inconsistent stochastic influence. These patterns match each method’s mechanism: MC-dropout modulates noise by epistemic uncertainty, supporting effective exploration, whereas AGN, though diversity can be increased via $\sigma$, injects less adaptive noise with variable impact across steps.

\begin{figure*}[t]
    \centering
        \begin{subfigure}[t]{0.42\textwidth}
        \includegraphics[width=\textwidth]{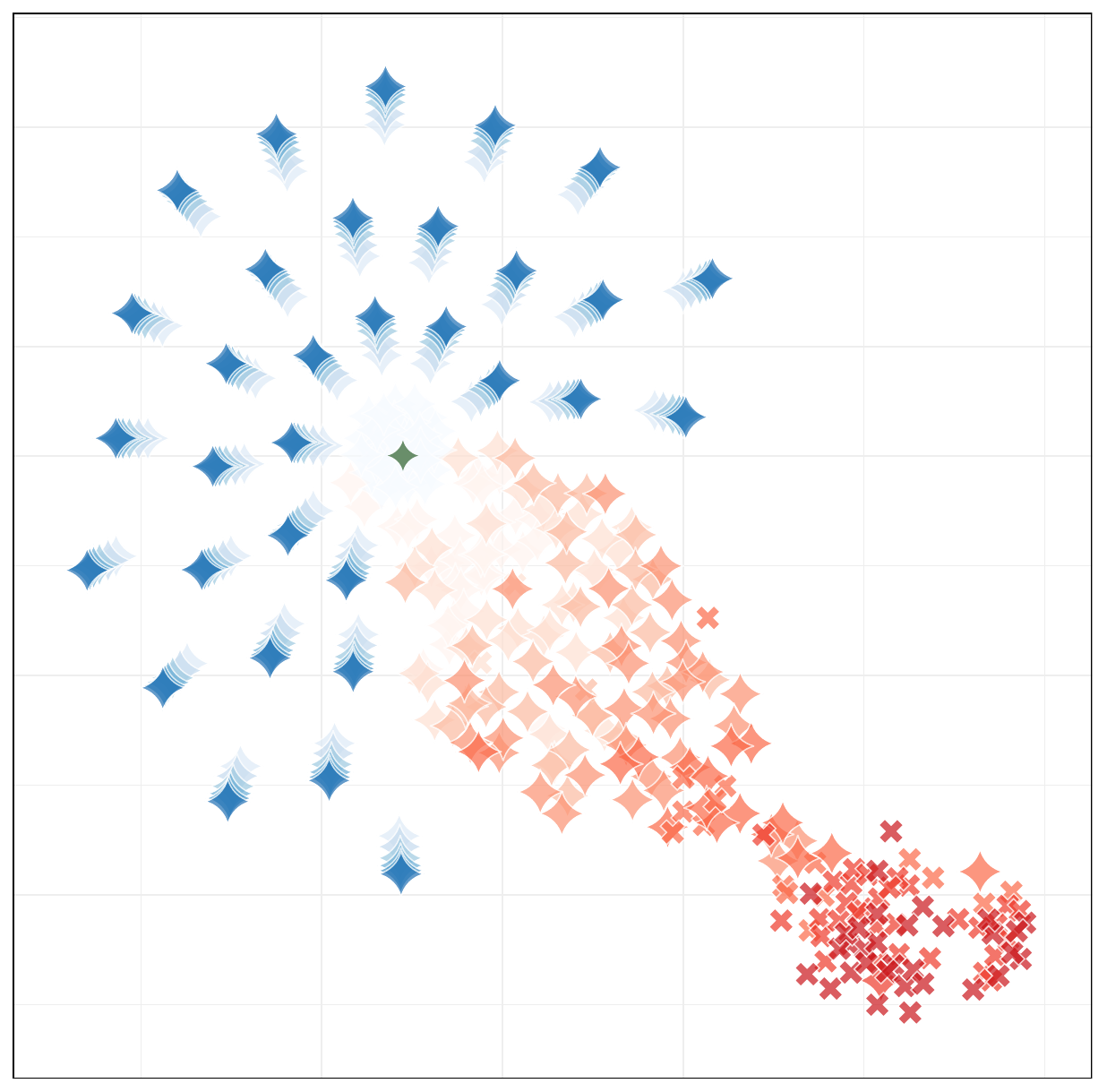}
        \caption{Easy}
        \label{fig:sub-easy}
    \end{subfigure}
    \begin{subfigure}[t]{0.42\textwidth}
        \includegraphics[width=\textwidth]{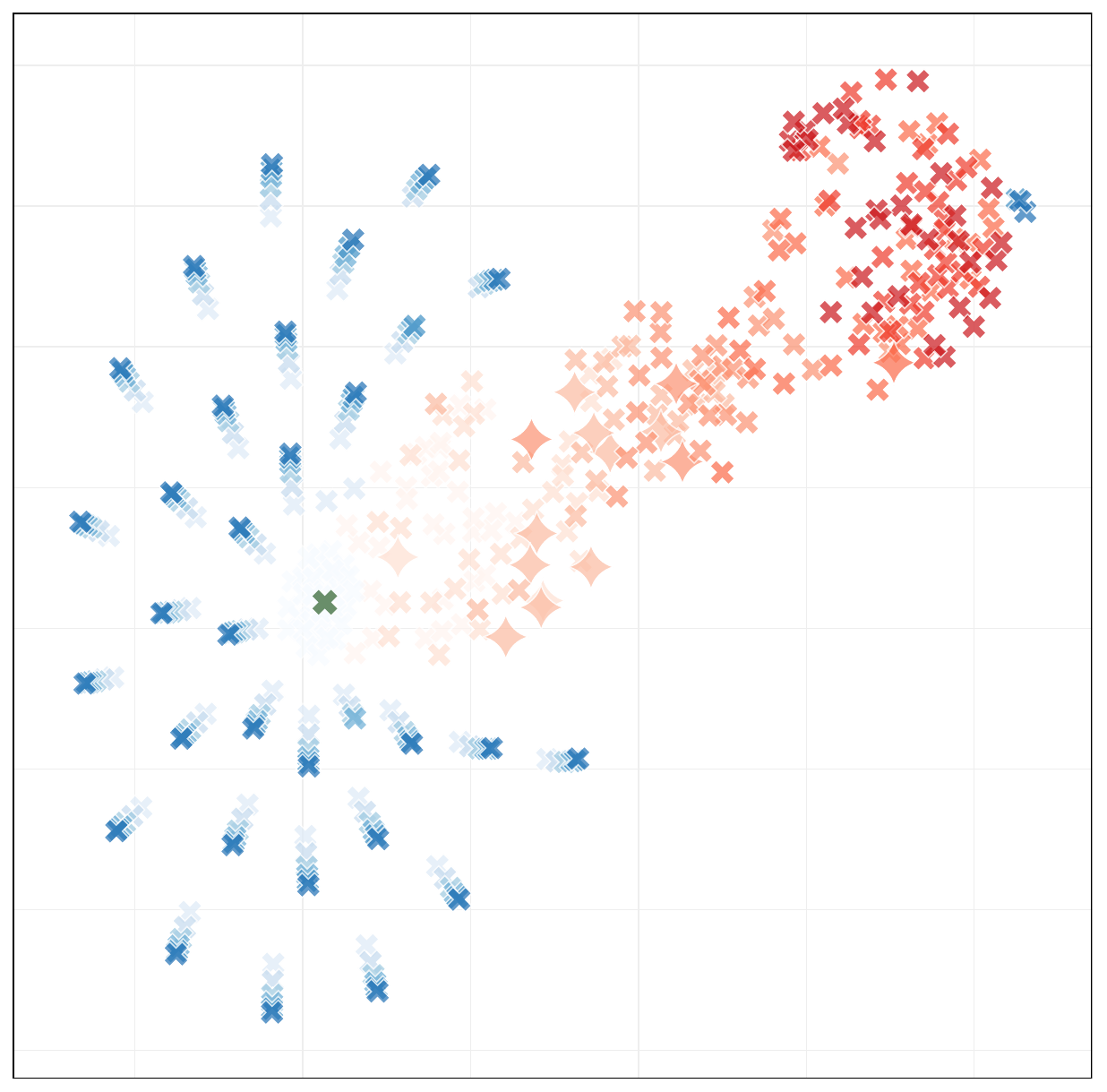}
        \caption{Hard}
        \label{fig:sub-hard}
    \end{subfigure}
    \begin{subfigure}[t]{0.13\textwidth}
        \includegraphics[width=\textwidth]{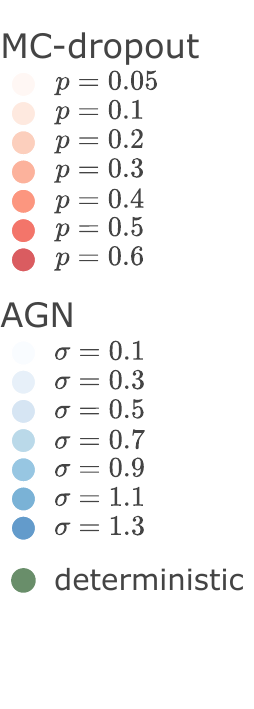}
    \end{subfigure}
    \caption{
    t-SNE visualization of latent thoughts sampled with different \textcolor{tsneCaptionRed}{dropout rates ($p$ from light to dark)} and \textcolor{tsneCaptionBlue}{Gaussian-noise scales ($\sigma$ from light to dark)}. The \textcolor{tsneCaptionGreen}{green marker} denotes the deterministic latent thought (no stochasticity). Diamonds ($\diamond$) indicate correct reasoning trajectories; crosses ($\bm{\times}$) indicate incorrect ones.  (\ref{fig:sub-easy}): an easy question. (\ref{fig:sub-hard}): a hard question. 
}
    \label{fig:tsne}
\end{figure*}

\begin{figure*}[thbp]
    \centering
        \begin{subfigure}[t]{0.325\textwidth}
        \includegraphics[width=\textwidth]{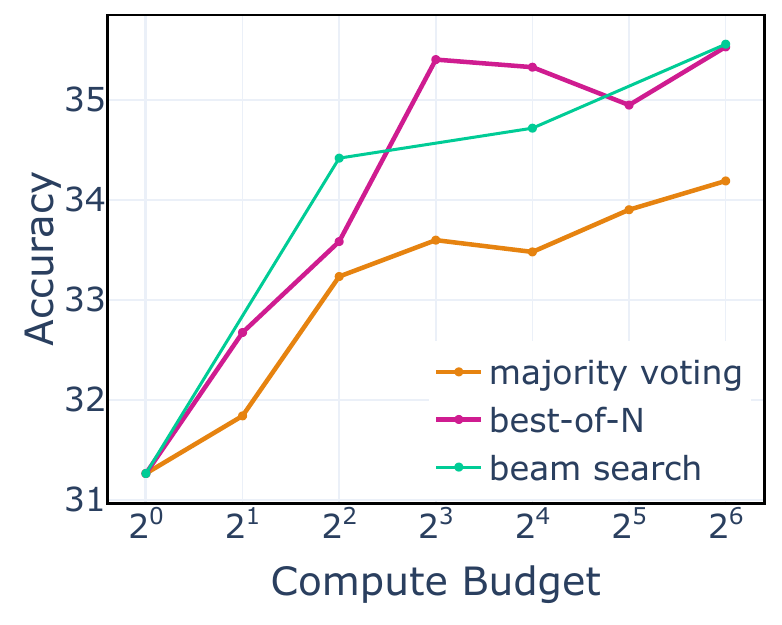}
        \caption{GSM-Test}
        \label{fig:tts-test}
    \end{subfigure}
    \hfill
    \begin{subfigure}[t]{0.325\textwidth}
        \includegraphics[width=\textwidth]{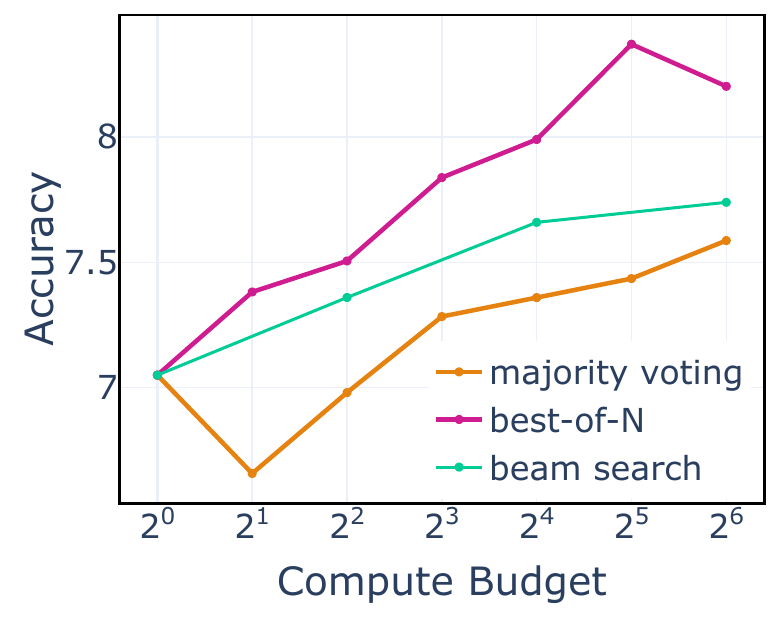}
        \caption{GSM-Hard}
        \label{fig:tts-hard}
    \end{subfigure}
    \hfill
    \begin{subfigure}[t]{0.325\textwidth}
        \includegraphics[width=\textwidth]{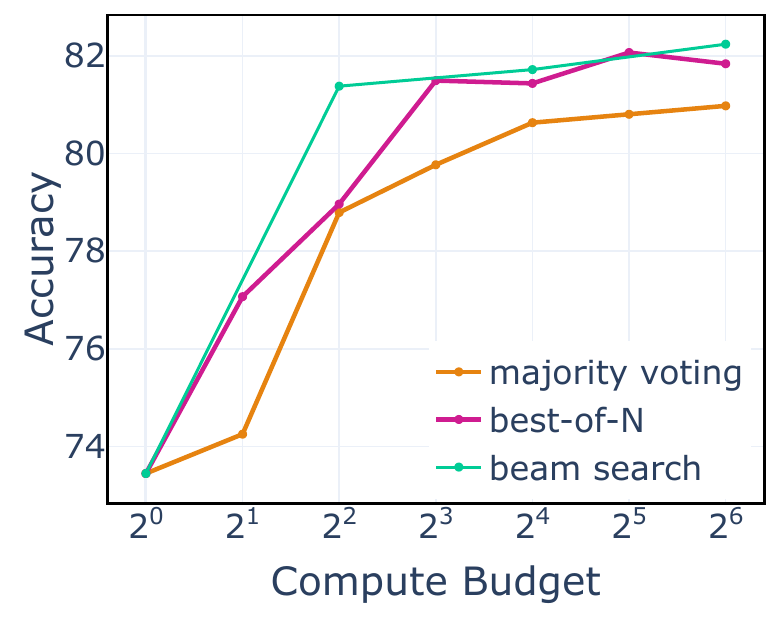}
        \caption{MultiArith}
        \label{fig:tts-multiarith}
    \end{subfigure}
    \caption{
{Test-time scaling results with majority voting, best-of-$N$, and beam search for latent reasoning.} Accuracy(\%) versus compute budget $N$ under three aggregation strategies on (\ref{fig:tts-test}) GSM8K-Test, (\ref{fig:tts-hard}) GSM8K-Hard, and (\ref{fig:tts-multiarith}) MultiArith. Results are reported as the mean over three runs.
}
    \label{fig:TTS}
\end{figure*}

\subsection{Visualization on Stochastic Latent Landscape}

To unveil how stochasticity reshapes the hidden landscape of reasoning, we cast sampled latent thoughts into a 2D stage via t-SNE (Figure~\ref{fig:tsne}).
What emerges is strikingly distinct:
Dropout produces a \emph{directional drift}—dense and contiguous along specific directions, whereas AGN yields an \emph{isotropic radial dispersion}, a “firework” pattern with broader area but lower local density.

These geometric signatures explain their different behaviors on easy and hard questions.
For easy cases (Figure~\ref{fig:sub-easy}), 
where correct regions lie near the deterministic latent, AGN maintains accuracy by keeping more probability mass around the center, whereas large $p$ dropout drifts away and degrades performance.
For hard cases (Figure~\ref{fig:sub-hard}), where correct regions are farther away, dropout’s larger displacement and denser exploration increase the chance of reaching the correct solution.
Ultimately, the geometry of exploration explains their complementary strengths: MC-dropout excels on harder tasks, whereas AGN maintains robustness even under strong stochasticity.

\section{Results of Aggregation}
\label{sec:aggregation-results}
\subsection{Main Results}

We report the TTS results using the three above-mentioned aggregation methods in Figure~\ref{fig:TTS}. 
\
First, accuracy increases monotonically with $N$ across all three datasets, demonstrating that latent reasoning can effectively scale with more inference compute. 
\
Second, both Best-of-$N$ and Beam Search consistently outperform Majority Voting, confirming that LatentRM can effectively distinguish promising reasoning trajectories.
\
Third, Beam Search performs comparably with Best-of-$N$ on GSM-Test and MultiArith but trails on GSM-Hard, suggesting that early-step score noise can cause premature pruning in more challenging problems. 
\
Finally, the gains are most pronounced on MultiArith, highlighting the generalization ability of the learned scorer across arithmetic reasoning patterns. 
\
Overall, LatentRM enables scalable and reliable aggregation, with Best-of-$N$ emerging as the most effective route for latent test-time scaling.

\begin{table}[t]
\centering
\small
\resizebox{.96\linewidth}{!}{
\begin{tabular}{lcc}
\toprule
\textbf{Variant} & \textbf{Test} & \textbf{Hard} \\
\midrule
Best-of-$8$ with LatentRM                            & \textbf{35.4} & \textbf{7.8} \\
\quad w/o contrastive (BCE)      & 33.5 & 7.4 \\
\quad w/o stochastic rollouts    & 30.7 & 6.0 \\
\quad random scalar head (untrained)            & 28.9 & 5.8 \\
\midrule
Majority Voting                  & 33.6 & 6.1 \\
\bottomrule
\end{tabular}
}
\caption{
\textbf{Ablation studies of LatentRM.}
We report accuracy (\%) on GSM-Test and GSM-Hard under Best-of-$N$ aggregation ($N=32$). 
Each variant isolates one design choice in LatentRM. 
}
\label{tab:latentrm-ablation}
\end{table}
\subsection{Ablation Studies}

To understand the contribution of each component in LatentRM, we conduct ablation studies under the Best-of-$N$ setting with $N=8$ on GSM-Test and GSM-Hard (\revised{Table~\ref{tab:latentrm-ablation}}). Each variant disables one design element of LatentRM while keeping all other configurations identical.
(1) Removing the step-wise contrastive loss (\emph{w/o contrastive}) causes a noticeable drop, showing that relative supervision among concurrent thoughts provides stronger learning signals than isolated binary labels. 
(2) Excluding stochastic rollouts (\emph{w/o stochastic rollouts}), where each thought is labeled only by the final trajectory correctness instead of intermediate estimates, leads to a further decline, highlighting the importance of Monte Carlo estimation for reliable thought-level annotation. 
(3) Using an untrained random scalar head (\emph{random head}) performs even worse—below Majority Voting—indicating that the gain stems from learned evaluation rather than architectural modification. 
Together, these findings verify that LatentRM effectively learns to assess latent trajectories and is crucial for successful aggregation in latent test-time scaling.

\subsection{More Analysis}
Due to space constraints, 
further extended analyses on \textit{Comparison of Latent and Explicit Reasoning with TTS}, \textit{Wall-Clock Comparison of Latent and Explicit Reasoning with TTS}, and \textit{Analysis on LatentRM under Variable-Length Setting} are provided in Appendix~\ref{sec:appendix-latent-vs-cot}, Appendix~\ref{sec:appendix-efficiency}, and Appendix~\ref{sec:appendix-variable-length}.

\section{Conclusion and Future Work}
\label{sec:conclusion}
This paper proposed a parallel test-time scaling framework for latent reasoning models, addressing the central challenges of sampling in continuous latent spaces and aggregating latent trajectories. By introducing two principled stochastic sampling methods—Monte Carlo Dropout and additive Gaussian noise—rooted in uncertainty theory, and developing a latent Reward Model trained with step-wise contrastive objective for effective trajectory aggregation, our approach enables scalable and robust parallel inference in the latent regime.
Extensive experiments and analyses demonstrate distinctive exploration behaviors, robust aggregation through latentRM, and consistent performance scaling across compute budgets, yielding new insights into test-time scaling for latent reasoning.

Building on this foundation, future research should explore two key directions: (1) integrating sampling and aggregation into a reinforcement learning framework to optimize latent trajectories through iterative feedback and reward shaping; and \revised{(2) investigating how to adapt ensemble techniques~\cite{bridging_gap_2024,breaking_ceiling_2024,hit_sweet_spot_2025} to latent reasoning, addressing the key challenge of applying token-based ensemble methods to continuous latent representations.}

\section*{Acknowledgment}

The work described in this paper was supported by the Research Grants Council of Hong Kong (\texttt{PolyU/15207122}, \texttt{PolyU/15213323}, \texttt{PolyU/15209724}, \texttt{PolyU/15205325}), the PolyU internal grants (BDWP), and the Special Fund for Taishan Scholar Project of Shandong Province.

\bibliographystyle{unsrtnat} 
\bibliography{zotero_references,references}

\appendix
\begin{paragraphrevised}
\section{Soft and Latent Reasoning}
\label{sec:appendix-soft-token-comparison}

Soft Thinking~\cite{llms_single_threaded_2025,SoftThinking} and Latent Reasoning~\cite{COCONUT, COLAR} represent two distinct paradigms for reasoning in large language models. Soft Thinking operates in the token probability space, where "soft tokens" are created by mixing token embeddings according to probability distributions over the vocabulary. This approach effectively retains the vocabulary structure while enabling continuous interpolation between discrete tokens~\cite{SoftThinking}. In contrast, Latent Reasoning works directly with pure latent vectors (hidden states) that are not constrained to be mixtures of token embeddings, operating in a more abstract and unconstrained continuous space~\cite{COCONUT}.

Wu et al.~\shortcite{wu_parallel_2025} reveal that Soft Thinking suffers from a greedy pitfall: LLMs predominantly rely on the highest-probability token, suppressing alternative reasoning paths. Their Stochastic Soft Thinking addresses this by introducing controlled randomness (via Gumbel-Softmax) in soft embedding composition. In contrast, our Parallel TTS for Latent Reasoning operates in pure latent vector space, unconstrained by token embeddings or vocabulary structure. We combine latent sampling (MC-Dropout or AGN) with a Latent Reward Model to enable parallel exploration and aggregation, discovering reasoning patterns unbound by vocabulary structure.
\end{paragraphrevised}

\section{Theoretical Perspectives on Sampling in Latent Space}
\label{sec:appendix-sampling}

We here briefly derive how MC-dropout can be interpreted as an approximate Bayesian estimator of \emph{epistemic} uncertainty, while additive Gaussian noise acts as a perturbation mechanism that simulates \emph{aleatoric} variability in latent representations.

\paragraph{MC-dropout} can be interpreted as an approximate Bayesian method, where stochastic forward passes approximate the posterior predictive distribution given the training dataset $D$~\cite{gal_dropout_2016}. Let the predictive distribution be
\[
p(y \mid x, D) = \int p(y \mid x, \omega)p(\omega \mid D)d\omega,
\]
where \(\omega\) denotes network weights and \(D\) the training data. Since the posterior \(p(\omega \mid D)\) is intractable, dropout approximates it with a variational distribution
\[
q(\omega): \quad W_i = M_i \cdot \mathrm{diag}(z_i), \quad z_{i,j} \sim \text{Bernoulli}(p_i).
\]
At inference, performing stochastic forward passes with dropout corresponds to sampling \(\omega^{(t)} \sim q(\omega)\) and evaluating
\[
{y}^{(n)} = f(x; \omega^{(n)}).
\]

The empirical mean and variance over \(N\) samples approximate the true Bayesian predictive distribution:
\[
\mathbb{E}[y^* \mid x] \approx \frac{1}{N} \sum_{t=1}^N f(x; \omega^{(n)}), 
\]
\[\mathrm{Var}[y^* \mid x] \approx \frac{1}{N} \sum_{t=1}^N f(x; \omega^{(n)})^2 - \Big(\mathbb{E}[y^* \mid x]\Big)^2 .\]
As \(T \to \infty\), this Monte Carlo procedure converges to the predictive distribution under the variational approximation.
Applying dropout at inference effectively samples from a distribution over model weights and creating an ensemble of predictions for each input. The variability across these predictions arises from the uncertainty in model parameters due to limited knowledge gained from training data $D$, this is defined as \emph{epistemic uncertainty}.

\paragraph{AGN} perturbs the latent representations directly, injecting variance proportional to the model’s local sensitivity. 
\[
x^* = x + \epsilon, \quad \epsilon \sim \mathcal{N}(0, \sigma^2 I).
\]

The predictive distribution then becomes
\[
p(y \mid x^*) = \int p(y \mid x + \epsilon) p(\epsilon)d\epsilon ,
\]
with \(p(\epsilon)\) an isotropic Gaussian prior. Expanding to first order (via Taylor approximation around \(x\)),
\[
f(x + \epsilon) \approx f(x) + J_f(x)\epsilon ,
\]
where \(J_f(x)\) is the Jacobian of the network output w.r.t. \(x\). Taking expectation yields
\[
\mathbb{E}[f(x+\epsilon)] \approx f(x),
\]
\[
\mathrm{Var}[f(x+\epsilon)] \approx J_f(x), \Sigma_\epsilon J_f(x)^\top \quad \Sigma_\epsilon = \sigma^2 I.
\]
AGN injects variance proportional to the network’s local sensitivity.
Since this variability is induced externally by perturbations to the latent inputs, it corresponds to \emph{aleatoric uncertainty}, i.e., randomness arising from inherent noise in the input space rather than uncertainty about the model itself.

\section{Derivation of Cumulative Scoring}
\label{sec:appendix-scoring}
Here we derive in detail why summing the logits over each candidate trajectory serves as a valid proxy for the relative quality of the generated thought sequence.
Consider $N$ reasoning trajectories, where each trajectory $n \in {1,\dots,N}$ produces a sequence of latent scores ${r_{t'}^{(n)}}_{t'=1}^t$.

At each step $t'$, the probability assigned to thought $n$ under the softmax over all candidates is
\[
p_{t'}^{(n)} = \frac{\exp(r_{t'}^{(n)})}{\sum_{n'=1}^N \exp(r_{t'}^{(n')})}.
\]

The log-probability is
\[
\log p_{t'}^{(n)} = r_{t'}^{(n)} - \log \sum_{n'=1}^N \exp(r_{t'}^{(n')}).
\]

The cumulative log probability over the first $t$ time steps for thought $n$ is:
\begin{align*}
 \log p_{1:t}^{(n)} & = \sum_{t'=1}^{t} \log p_{t'}^{(n)} \\
& =  \sum_{t'=1}^{t} r_{t'}^{(n)} - \sum_{t'=1}^{t} \log \sum_{n'=1}^N \exp(r_{t'}^{(n')}).
\end{align*}

The second term
\(
 \sum_{t'=1}^{t} \log \sum_{n'=1}^N \exp(r_{t'}^{(n')})
\)
is identical for all trajectories at step $t$. Therefore, when comparing trajectories, the relative ordering of $\log p_{1:t}^{(n)}$ depends only on the first term $\sum_{t'=1}^{t} r_{t'}^{(n)}$.

\section{Inference Procedures for Aggregation}
\label{sec:appendix-aggregation-algo}

We detail the inference algorithms used to aggregate multiple latent reasoning trajectories, corresponding to the three strategies evaluated in Section~\ref{sec:aggregation-results}. All procedures operate on $N$ sampled latent trajectories $\{h^{(n)}_{1:T}\}_{n=1}^{N}$ obtained via the stochastic sampling methods introduced in Section~\ref{sec:sampling}. Each latent thought $h^{(n)}_t$ is scored by the Latent Reward Model (LatentRM) as $r^{(n)}_t = g_\phi(x, h^{(n)}_{1:t})$.

\paragraph{Best-of-$N$.}
Each trajectory is evaluated independently using the cumulative latent reward 
\[
R^{(n)} = \sum_{t=1}^{T} r^{(n)}_t .
\]
The trajectory with the highest total reward is selected for final decoding:
\[
\bm{h}^{*}_{1:T} = \arg\max_{n} R^{(n)}.
\]
The final answer token sequence $y^{*}$ is then produced by the reasoning model conditioned on $h^{*}_{1:T}$:
\[
y^{*} = {f}_{\bm{\theta}}(\bm{x}, \bm{h}^{*}_{1:T}).
\]
This mirrors best-of-$N$ decoding in token-based TTS, but replaces token log-likelihoods with learned latent rewards.

\paragraph{Beam Search.}
We further employ a beam search guided by LatentRM to explore high-reward reasoning paths.  
At step $t$, the model expands each partial latent trajectory in the beam $\mathcal{B}_{t-1}$ by one autoregressive step, producing $K$ candidate extensions via stochastic sampling:
\[
\tilde{\bm{h}}^{(k)}_{1:t} = [{\bm{h}}^{(b)}_{1:t-1}, \hat{\bm{h}}^{(k)}_t], \quad \hat{\bm{h}}^{(k)}_t \sim f_{\bm{\theta}}({\bm{h}}^{(b)}_{1:t-1}, \bm{x}),
\]
where $b$ indexes a beam element.  
LatentRM assigns scores $r^{(k)}_t$ to all candidates, and their cumulative rewards are updated:
\[
R^{(k)}_t = R^{(b)}_{t-1} + r^{(k)}_t.
\]
The top-$B$ candidates by cumulative reward are retained as the next beam:
\[
\mathcal{B}_t = \text{TopB}(\{\tilde{h}^{(k)}_{1:t}\}, R^{(k)}_t).
\]
Decoding terminates when all beams emit the end-of-thinking token or reach the maximum latent step $T$. The best final trajectory is selected by its cumulative reward.  
In experiments, we set $B = \sqrt{N}$ to match compute cost with best-of-$N$ as discussed in Section~\ref{subsec:exp-aggregation}.

\paragraph{Majority Voting.}
As a non-parametric baseline, each latent trajectory is independently decoded into an answer $a^{(n)}$. The final output is the most frequent answer among the $N$ candidates:
\[
y^{*} = \text{mode}(\{a^{(n)}\}_{n=1}^{N}).
\]
This strategy disregards latent scores and serves to isolate the benefit of learned aggregation via LatentRM.

\paragraph{Implementation Details.}
All aggregation procedures are executed under identical compute budgets. For best-of-$N$ and majority voting, $N$ full trajectories are sampled; for beam search, the beam size $B=\sqrt{N}$ and per-step expansion $K=B$ ensure comparable total forward passes. Cumulative rewards $\sum_t r^{(n)}_t$ are pre-normalized by trajectory length to prevent bias toward longer reasoning chains.

\

\begin{paragraphrevised}
\section{Additional Results on Harder Benchmarks}
\label{sec:appendix-harder-benchmarks}
We further evaluate our method on harder benchmarks, GPQA and AIME. As shown in Figure~\ref{fig:sampling-harder}, our method induces consistent performance gains even on these harder benchmarks, which demonstrates the general effectiveness of the proposed approach. However, absolute performance remains limited. We attribute to two factors: (a) small model size (e.g., GPT-2-124M, Llama-3.2-1B), and (b) the inherent challenges of the "latent reasoning" framework, which is still an evolving paradigm and struggle on advanced mathematics (AIME) and Ph.D.-level benchmark (GPQA).

\begin{figure}[t]
    \centering
    \begin{subfigure}[b]{\linewidth}
        \includegraphics[width=\linewidth]{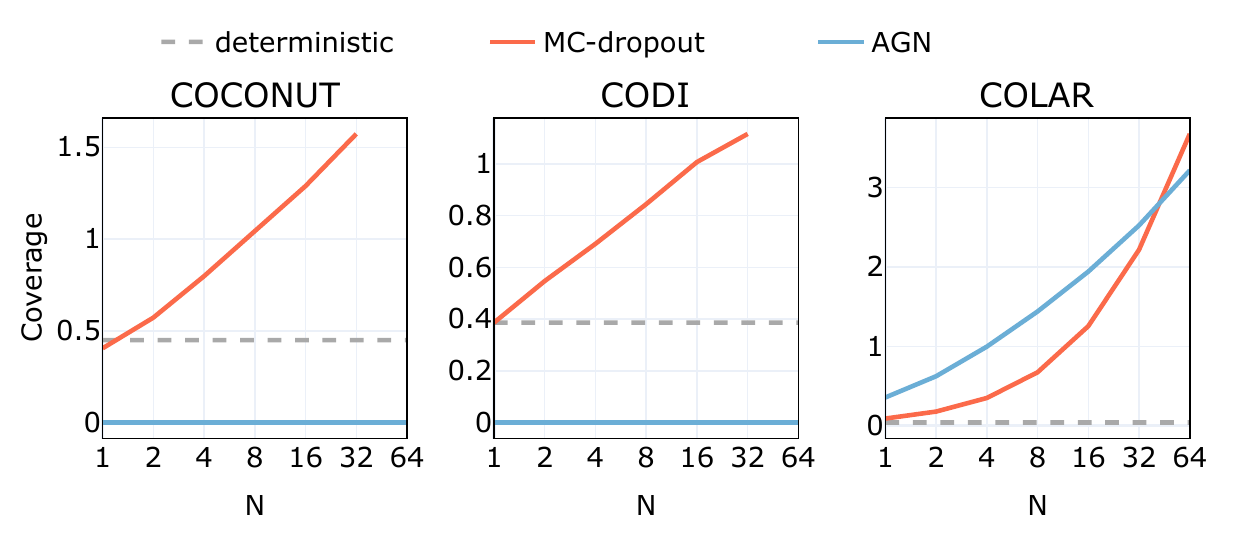}
        \caption{GPQA}
        \label{fig:sampling:gpqa}
    \end{subfigure}
    \hfill
    \begin{subfigure}[b]{\linewidth}
        \includegraphics[width=\linewidth]{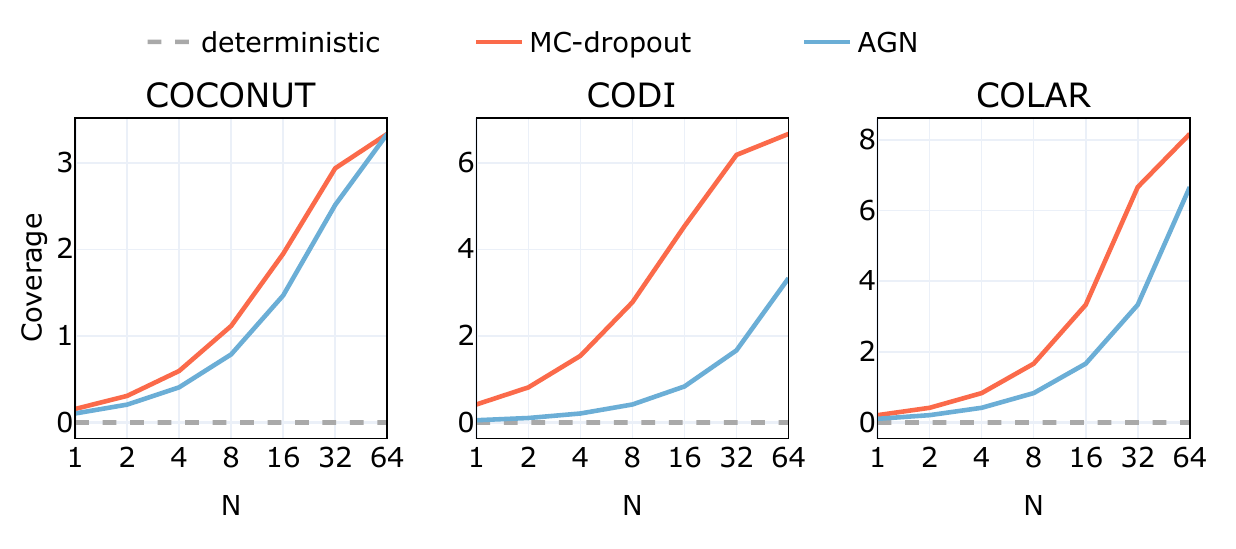}
        \caption{AIME}
        \label{fig:sampling:aime}
    \end{subfigure}
    \caption{\revised{Scaling analysis on harder benchmarks (GPQA and AIME).}}
    \label{fig:sampling-harder}
\end{figure}

\section{More Analysis}

\subsection{Comparison of Latent and Explicit Reasoning with TTS}
\label{sec:appendix-latent-vs-cot}
To investigate the synergy between latent reasoning and Test-Time Scaling (TTS), we compare \textsc{No-CoT} (direct answering), explicit \textsc{CoT}, and \textsc{COCONUT} (latent reasoning), evaluating each with and without parallel TTS ($N=4$). We measure computational overhead by the number of decoding steps relative to the \textsc{No-CoT} baseline.

\begin{table}[t]
    \centering
    \setlength{\tabcolsep}{10pt} 
    \resizebox{\linewidth}{!}{
    \begin{tabular}{l ccc}
    \toprule
    \textbf{Method} & \textbf{GSM-Test} & \textbf{GSM-Hard} & \textbf{Compute} \\
    \midrule
    \rowcolor[gray]{0.95} NO-COT & & & \\
    deterministic & 16.5 & 4.3 & $\times$1 \\
    Coverage@4 & 22.3 & 5.5 & $\times$4 \\
    \midrule
    \rowcolor[gray]{0.95} COT & & & \\
    deterministic & 42.9 & 9.8 & $\times$30$^{\dagger}$ \\
    Coverage@4 & 50.7 & 11.6 & $\times$120$^{\dagger}$ \\
    \midrule
    \rowcolor[gray]{0.95} COCONUT & & & \\
    deterministic & 34.0 & 7.0 & \textbf{$\times$6} \\
    Coverage@4 & \textbf{46.7} & \textbf{9.5} & $\times$24 \\
    Best-of-4 & 34.5 & 7.5 & $\times$24 \\
    \bottomrule
    \multicolumn{4}{l}{\footnotesize $^{\dagger}$ Average compute across Test and Hard subsets.}
    \end{tabular}
    }
    \caption{\revised{Performance and computational efficiency on GPT-2. Latent reasoning (COCONUT) achieves a superior balance between accuracy and computation compared to explicit CoT.}}
    \label{tab:latent-vs-cot}
\end{table}

As summarized in Table~\ref{tab:latent-vs-cot}, latent reasoning combined with TTS demonstrates superior efficiency, achieving competitive performance with significantly lower computational costs than explicit \textsc{CoT}. Specifically, \textsc{COCONUT} with TTS outperforms the explicit \textsc{CoT} baseline on the same backbone while requiring fewer decoding steps. Furthermore, while a performance gap persists between the Best-of-$N$ with LatentRM and the Coverage@$N$ upper bound, integrating LatentRM provides consistent gains. These results suggest that latent reasoning, augmented by TTS, preserves the efficiency advantages of continuous-space processing while matching or exceeding the accuracy of discrete reasoning.

\subsection{Wall-Clock Comparison of Latent and Explicit Reasoning with TTS}
\label{sec:appendix-efficiency}
To complement the decoding-step comparison in Table~\ref{tab:latent-vs-cot}, we further contrast \textsc{COCONUT} (latent reasoning) and explicit \textsc{CoT} under parallel Test-Time Scaling (TTS), using the settings summarized below. We measure end-to-end wall-clock latency in seconds per question on a single NVIDIA H100 GPU, and report sampling time, Best-of-$N$ aggregation time, and total latency.

\begin{table}[t]
\centering
\small
\setlength{\tabcolsep}{4pt}
\label{tab:efficiency}
\resizebox{.9\linewidth}{!}{
\begin{tabularx}{\columnwidth}{@{} >{\raggedright\arraybackslash}l *{4}{>{\centering\arraybackslash}X} @{}}
\toprule
\textbf{Setting} & \textbf{Sampling} & \makecell{\textbf{Agg.}} & \textbf{Total} \\
\midrule
\rowcolor[gray]{0.95} \textit{$N$=1} & & & \\
COCONUT & 0.050 & --- & \textbf{0.050}  \\
COT & --- & --- & 0.400  \\
\midrule
\rowcolor[gray]{0.95} \textit{$N$=32} & & & \\
COCONUT  & 1.322 & 0.386 & \textbf{1.708} \\
COT & --- & --- & 9.802 \\
\bottomrule
\end{tabularx}
}
\caption{Wall-clock latency comparison. Aggregation remains efficient even as $N$ increases.}
\end{table}

As summarized in Table~\ref{tab:efficiency}, \textsc{COCONUT} with TTS retains a favorable wall-clock profile relative to explicit \textsc{CoT} at matched $N$: aggregation contributes only a modest fraction of total time (on the order of 20\% at $N{=}32$), so parallel sampling rather than reranking dominates cost. The gap widens as $N$ increases, consistent with short latent trajectories ($T{=}6$ thoughts) versus long explicit chain-of-thought rollouts, together with sequence-level aggregation.

\subsection{Analysis on LatentRM under Variable-Length Setting}
\label{sec:appendix-variable-length}
\begin{table}[t]
    \centering
    \small
    \begin{tabular}{lc}
    \toprule
    \textbf{Method} & \textbf{Accuracy (\%)} \\
    \midrule
    Deterministic & 32.44 \\
    Best-of-4 & 33.81 \\
    Best-of-8 & 33.95 \\
    Best-of-16 & 34.04 \\
    \bottomrule
    \end{tabular}
    \caption{\revised{Performance with variable-length latent reasoning paths on GSM8K-Test.}}
    \label{tab:variable-length}
    \end{table}
Latent reasoning models may generate trajectories of varying lengths, we thus conduct an evaluation of LatentRM in a variable-length setting.
We modify the COCONUT paradigm by introducing stochasticity into the trajectory generation process, sampling the number of reasoning steps uniformly from 4 to 6, instead of fixing the trajectory length to 6.
To mitigate potential length bias in trajectory selection, we utilize the mean step-wise reward, defined as the total reward divided by the step count, to normalize scores. 
Experimental results in Table~\ref{tab:variable-length} indicate that best-of-$N$ reranking based on average reward consistently outperforms the deterministic baseline, suggesting that LatentRM can effectively handle variable-length latent reasoning.

\end{paragraphrevised}
\section{Training Configuration for LatentRM}
\label{sec:appendix-training-latentRM}
We train a Latent Reward Model (LatentRM) for COCONUT using the step-wise contrastive learning objective introduced in Section~\ref{subsec:methods-aggregation}. model initialized from COOCNUT checkpoint, a GPT-2 backbone with 124 million parameters~\cite{radford_language_nodate}, with architecture modifications detailed therein.

\paragraph{Training data.}
The training data consists of samples generated based on \textbf{GSM8K} training set (385K examples). 
For each input problem, we sample $N=8$ reasoning trajectories via MC-dropout with dropout probability $p=0.2$ from COCONUT, where $p$ was tuned for optimal performance as mentioned in Section~\ref{subsec:sampling-main-results}.

\paragraph{Labeling.}
For each intermediate reasoning step within a trajectory, we perform $M = 128$ stochastic rollouts to empirically estimate the correctness of that step.
trajectories that are either too easy (\emph{i.e.,} all trajectories are correct) or too difficult (i.e., all trajectories are incorrect) are excluded to improve training stability and focus on informative examples.

\paragraph{Training configuration.}

\begin{itemize}
    \setlength{\itemsep}{0pt}
    \setlength{\parskip}{0pt}
    \setlength{\parsep}{0pt}
    \item Batch size: 2048
    \item Optimizer: Paged AdamW
    \item Learning rate: $1 \times 10^{-5}$
    \item Epochs: 10
    \item Hardware: 2 × NVIDIA H100 GPUs
\end{itemize}

\begin{paragraphrevised}
\section{Algorithmic Specification of Latent Sampling Procedures}
This section presents the pseudocode for the stochastic sampling procedures used in latent test-time scaling. We detail the inference-time execution of MC-Dropout and AGN for generating multiple latent reasoning trajectories.

\label{sec:appendix-sampling-algo}
\begin{algorithm}[H]
\caption{Sampling with MC-Dropout}
\label{alg:mc_dropout}
\begin{algorithmic}[1]
\small
\Require Input sequence $x$; model $f_\theta^{\text{latent}}$; dropout rate $p$; sample count $N$; stop-thinking condition STOP$(\cdot)$.
\State \textbf{Prefill:} Encode $x$ and initialize kv-cache.

\For{$n = 1$ to $N$}

    \State $t \gets 0$    
    \While{not $\text{STOP}(h^{(n)}_{1:t})$} 
            \State $m_t^{(n)} \sim \text{Bernoulli}(1-p)$ \Comment{\small Sample dropout mask}
        \State $\theta_t^{(n)} \gets \theta \odot m_t^{(n)}$ \Comment{\small Masked Subnetwork}
        \State $h^{(n)}_{t+1} \gets f_{\bm{\theta}^{(n)}}(x, h^{(n)}_{1:t})$ 
        \Comment{Forward pass}
        \State $t \gets t + 1$
    \EndWhile
    \State Store trajectory $\mathbf{h}^{(n)} \gets [h^{(n)}_1, \ldots, h^{(n)}_t]$
    \State Generate final answers $y^{(n)} \gets f_\theta (x, \mathbf{h}^{(n)})$
\EndFor

\end{algorithmic}
\end{algorithm}

\begin{algorithm}[H]
\caption{Sampling with AGN}
\label{alg:agn}
\begin{algorithmic}[1]
\small
\Require Input sequence $x$; model $f_\theta^{\text{latent}}$; noise scale $\sigma$; sample count $N$; stop-thinking condition STOP$(\cdot)$.
\State \textbf{Prefill:} Encode $x$ and initialize kv-cache.
\For{$n = 1$ to $N$}
    \State $t \gets 0$
    \State Initialize latent state $h^{(n)}_{1:0} \gets \emptyset$
    \While{not $\text{STOP}(h^{(n)}_{1:t})$}
        \State $\epsilon_t^{(n)} \sim \mathcal{N}(0, \sigma^2 I)$ \Comment{\small Sample Gaussian noise}
        \State $\tilde{h}_t^{(n)} \gets h_t^{(n)} + \epsilon_t^{(n)}$ \Comment{\small Perturb latent thought}
        \State $h^{(n)}_{t+1} \gets f_\theta^{\text{latent}}(x, \tilde{h}^{(n)}_{1:t})$ \Comment{\small Forward pass}
        \State $t \gets t + 1$
    \EndWhile
    \State Store trajectory $\mathbf{h}^{(n)} \gets [h^{(n)}_1, \ldots, h^{(n)}_t]$
    \State Generate final answers $y^{(n)} \gets f_\theta (x, \mathbf{h}^{(n)})$
\EndFor

\end{algorithmic}
\end{algorithm}

\end{paragraphrevised}

\end{document}